%% file: main.tex
\documentclass[10pt,twocolumn,letterpaper]{article}

\usepackage{cvpr}              

\usepackage{times}
\usepackage{epsfig}
\usepackage{graphicx}
\usepackage{amsmath}
\usepackage{amssymb}

\definecolor{cvprblue}{rgb}{0.21,0.49,0.74}
\usepackage[pagebackref,breaklinks,colorlinks,allcolors=cvprblue]{hyperref}

\def\paperID{17} 
\def\confName{CVPR}
\def\confYear{2025}

\definecolor{grayhighlight}{RGB}{213,229,255}

\newsavebox\CBox

\begin{document}

\title{Efficient Image Denoising on Smartphone GPUs: \\ Mobile AI 2026 Challenge Report}

\author{
Andrey Ignatov \and Georgy Perevozchikov \and Radu Timofte \and
HaoDong Wu \and Desheng Cai \and
Nikolay Falaleev \and
Faraz Kayani \and Sarmad Kayani \and Asad Ahmed \and Dmitry Ignatov \and
Kihwan Yoon \and Ganzorig Gankhuyag \and
Yoonsuk Kang \and The Van Le \and Jin Young Lee \and
Sunzhe Yang \and Qiuhai Yan \and Kang Chen \and Tingting Wang \and Faming Fang
}

\maketitle

\begin{abstract}
	
Image denoising is a fundamental problem in mobile photography. Despite a large number of solutions proposed for this task, the majority of them are trained on synthetically produced noisy images, and are additionally too computationally heavy for processing full-resolution images on mobile devices. This creates a need for efficient deep learning-based models that are not only performant but can also be deployed on constrained mobile hardware. In this Mobile AI challenge, we address this problem and propose the participants to design fast solutions that can denoise Full HD resolution photos while running on popular smartphone GPUs from Qualcomm and Arm. For this, the participants were provided with a large-scale dataset consisting of noisy-clean image pairs captured in the wild. The runtime of all models was evaluated on the Snapdragon 8 Elite Gen 5 and Dimensity 9500 chipsets with the Adreno and Mali GPUs, respectively. Top solutions proposed in this challenge are able to perform high-quality image denoising under 100 milliseconds on both platforms while reaching a PSNR score of over 37 dB. A comprehensive description of the models developed in the challenge is provided in this paper.

\end{abstract}
{\let\thefootnote\relax\footnotetext{%
\hspace{-5mm}$^*$Andrey Ignatov, Georgy Perevozchikov and Radu Timofte are the organizers of the Mobile AI 2026 challenge \textit{(andrey@vision.ee.ethz.ch, georgii.perevozchikov@uni-wuerzburg.de, radu.timofte@uni-wuerzburg. de)}. The other authors participated in the challenge. \\ Appendix \ref{sec:apd:team} contains the authors' team names and affiliations.\\ \\ Mobile AI 2026 Workshop website: \\ \url{https://ai-benchmark.com/workshops/mai/2026/}
}}

\section{Introduction}

\begin{figure*}[t!]
	\centering
	\setlength{\tabcolsep}{1pt}
	\resizebox{\linewidth}{!}
	{
		\includegraphics[width=0.5\linewidth]{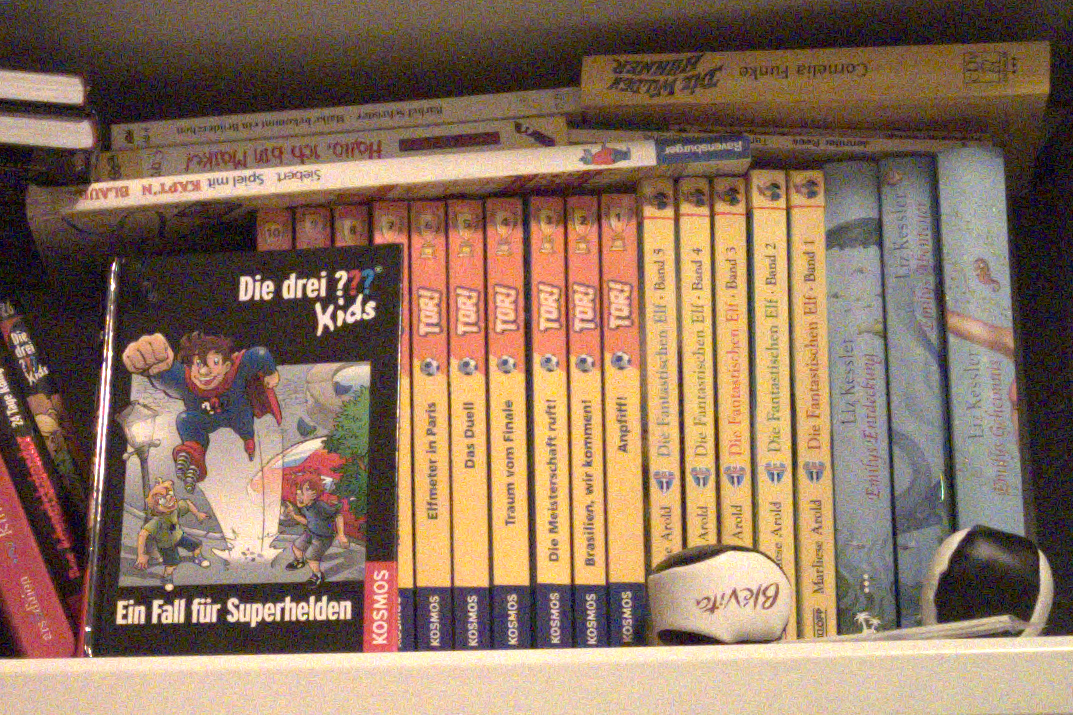} \hspace{2mm}
		\includegraphics[width=0.5\linewidth]{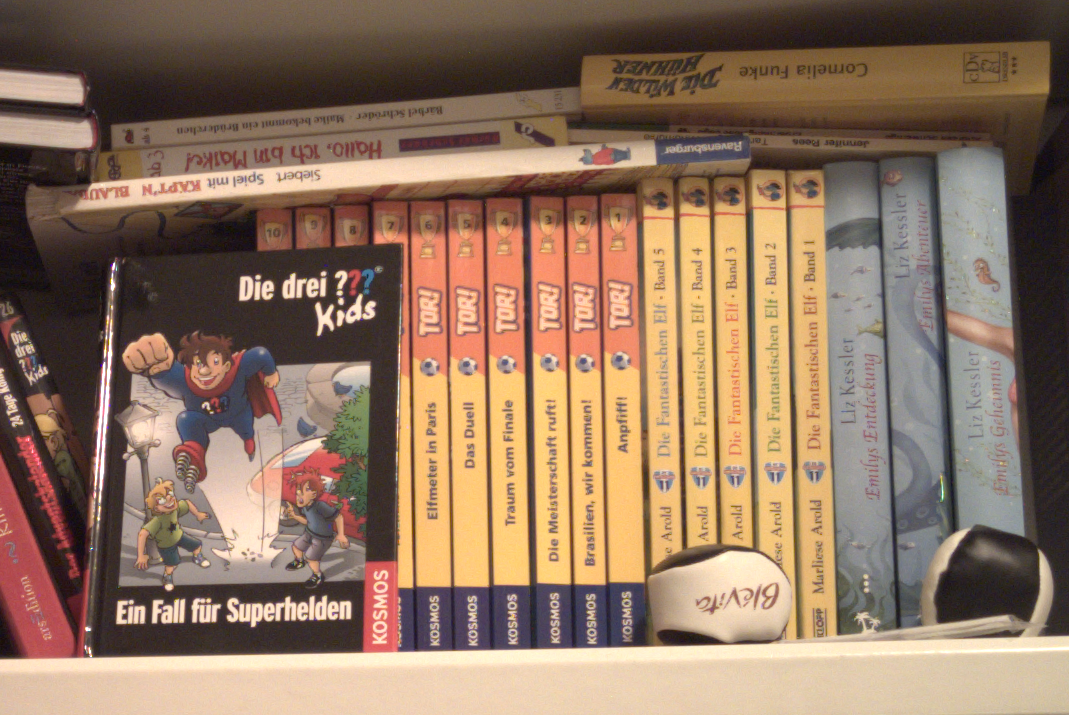}
	}
	\vspace{0cm}
	\caption{Sample crops from the original and denoised images from the collected dataset. Best zoomed on screen.}
	\label{fig:example_photos}
	\vspace{-0.2cm}
\end{figure*}

Despite recent advancements in mobile imaging sensor technology, image denoising remains a significant challenge in the processing of mobile photo and video data. Physical and design constraints limit the ability to increase sensor size in mobile devices~--- when combined with the trend toward higher sensor resolutions and reduced pixel dimensions, these factors contribute to elevated noise levels, particularly under low-light conditions. To address this issue, numerous classical denoising methods have been proposed in the past~\cite{kuan1985adaptive,liu2007automatic,dabov2007image,buades2005non,buades2005review,motwani2004survey,mihcak1999low,starck2002curvelet}. More recently, deep learning-based approaches have demonstrated substantially improved quantitative performance~\cite{gu2019brief,tai2017memnet,zhang2018ffdnet,zhang2017beyond,abdelhamed2020ntire,abdelhamed2019ntire,liang2021swinir,zamir2022restormer,chen2022simple,izadi2023image,li2023ntire,sun2025tenth,guo2025mambairv2}. However, despite achieving high fidelity metrics, many of these methods rely on synthetically generated training and validation datasets~\cite{tai2017memnet,zhang2018ffdnet,zhang2017beyond,liang2021swinir,zamir2022restormer,chen2022simple,li2023ntire,sun2025tenth,guo2025mambairv2}, or are evaluated on limited collections of indoor imagery~\cite{abdelhamed2018high,abdelhamed2020ntire,abdelhamed2019ntire,izadi2023image}, thereby restricting their generalizability to real-world noisy camera data. Another primary limitation of these approaches is that they are typically focused on optimizing the restoration quality, while neglecting the computational efficiency required for mobile platforms, where resource constraints are a critical consideration. In this challenge, we directly address this problem and propose the participants to use a real-world dataset and design efficient image denoising models that can run on recent smartphone GPUs.

Designing deep learning models that run efficiently on mobile devices~--- especially across different AI accelerators like NPUs~--- is a complex challenge. It demands careful analysis of hardware constraints such as limited operator support, restricted memory capacity and bandwidth, reduced accuracy for quantized models and specific driver behavior of NPUs. Even when certain layers or operations are technically supported, they may be executed inefficiently due to constraints related to NPU hardware architecture or its driver implementation, and thus create performance bottlenecks during inference. A striking example here would be the Google Tensor NPU present in all recent Google Pixel smartphones that, despite having quite high theoretical peak performance, is unable to efficiently accelerate the vast majority of even simplistic CNN-based models~\cite{ignatov2025MAISR}. As a result, achieving low latency and energy efficiency often requires partial or complete redesigns of model architectures. For a more detailed discussion of these challenges and the performance of modern mobile AI accelerators we refer to~\cite{ignatov2019ai,ignatov2018ai}. Overall, all these limitations require various adaptations of the standard deep learning models for their efficient deployment on mobile NPUs, including conventional optimizations like 16-bit / 8-bit~\cite{chiang2020deploying,jain2019trained,jacob2018quantization,yang2019quantization,li2021mqbench,nagel2021white,bondarenko2021understanding,siddegowda2022neural,tan2024mobilequant,lin2025awq} and low-bit~\cite{cai2020zeroq,uhlich2019mixed,banner2019post,ignatov2020controlling,liu2018bi,zhuo2022empirical,bhalgat2020lsq+,kim2023memory,dettmers2023case,bondarenko2024low,li2025memory} quantization, network pruning and compression~\cite{chiang2020deploying,li2019learning,
liu2019metapruning,obukhov2020t,kim2021pqk,hawks2021ps,zheng2022model,xu2023sparsebytenn,kuzmin2023pruning,wang2024towards}, NPU-specific adaptations, platform-aware neural architecture search~\cite{howard2019searching,tan2019mnasnet,wu2019fbnet,
wan2020fbnetv2,mills2021profiling,chu2021discovering,zhan2021achieving,wu2022compiler,chitty2022neural,yoshihama2023hardware}, \etc.

In this Mobile AI Challenge, the participants were proposed to develop efficient deep learning models that can denoise Full HD resolution images directly on smartphone GPUs. The participating teams were provided with a large-scale image denoising dataset obtained with a recent Sony mobile camera sensor capturing photos in the burst mode. The obtained for each scene images were averaged to get a clean photo, and the resulting noisy-clean image pairs were used to train an end-to-end deep learning solution for this task. The runtime of the models was evaluated on mobile GPUs present in the Snapdragon 8 Elite Gen 5 and the Dimensity 9500. Each submission was evaluated based on both image fidelity and runtime efficiency, thus balancing reconstruction quality with model efficiency. All solutions developed during this challenge are fully compatible with the LiteRT~\cite{LiteRT2025} (former TensorFlow Lite) framework, enabling deployment on a variety of Linux and Android-based IoT platforms, smartphones, and edge devices.

\smallskip


This challenge is a part of the \textit{MAI 2026 Workshop and Challenges} consisting of the following competitions:



\begin{itemize}
\item 4K Image Super-Resolution on Mobile NPUs~\cite{ignatov2025sr4k}
\item Real-Time Video Super-Resolution on Smartphone GPUs~\cite{ignatov2025srvideo}
\item Efficient Image Denoising on Smartphone GPUs
\item RGB Photo Enhancement on Smartphone GPUs~\cite{ignatov2025rgb}
\item Learned Smartphone ISP on Mobile GPUs~\cite{ignatov2025isp}
\item Bokeh Effect Rendering on Smartphone GPUs~\cite{ignatov2025bokeh}
\item Efficient LLMs for On-Device Deployment
\item Adapting Stable Diffusion for Efficient On-Device Inference
\end{itemize}

\normalsize


\noindent The results and solutions obtained in the previous \textit{Mobile AI Challenges} are described in the following papers:


\begin{itemize}
\item Image Super-Resolution on Mobile NPUs~\cite{ignatov2021real,ignatov2022efficient,ignatov2025MAISR}
\item Learned ISP on Mobile NPUs and GPUs~\cite{ignatov2021learned,ignatov2022learned,ignatov2025MAIRAW}
\item Video Super-Resolution on Mobile GPUs~\cite{romero2021real,ignatov2022power}
\item Image Denoising on Mobile GPUs~\cite{ignatov2021fastDenoising}
\item Bokeh Effect Rendering on Mobile GPUs~\cite{ignatov2022realistic}
\item Camera Scene Detection on Smartphones~\cite{ignatov2021fastSceneDetection,pouget2021fast}
\item RGB Photo Enhancement on Mobile GPUs~\cite{ignatov2025MAIRGB}
\item Depth Estimation on Mobile Devices~\cite{ignatov2021fastDepth,ignatov2022efficient}
\end{itemize}

\normalsize

\begin{figure*}[t!]
\centering
\setlength{\tabcolsep}{1pt}
\resizebox{0.96\linewidth}{!}
{
\includegraphics[width=1.0\linewidth]{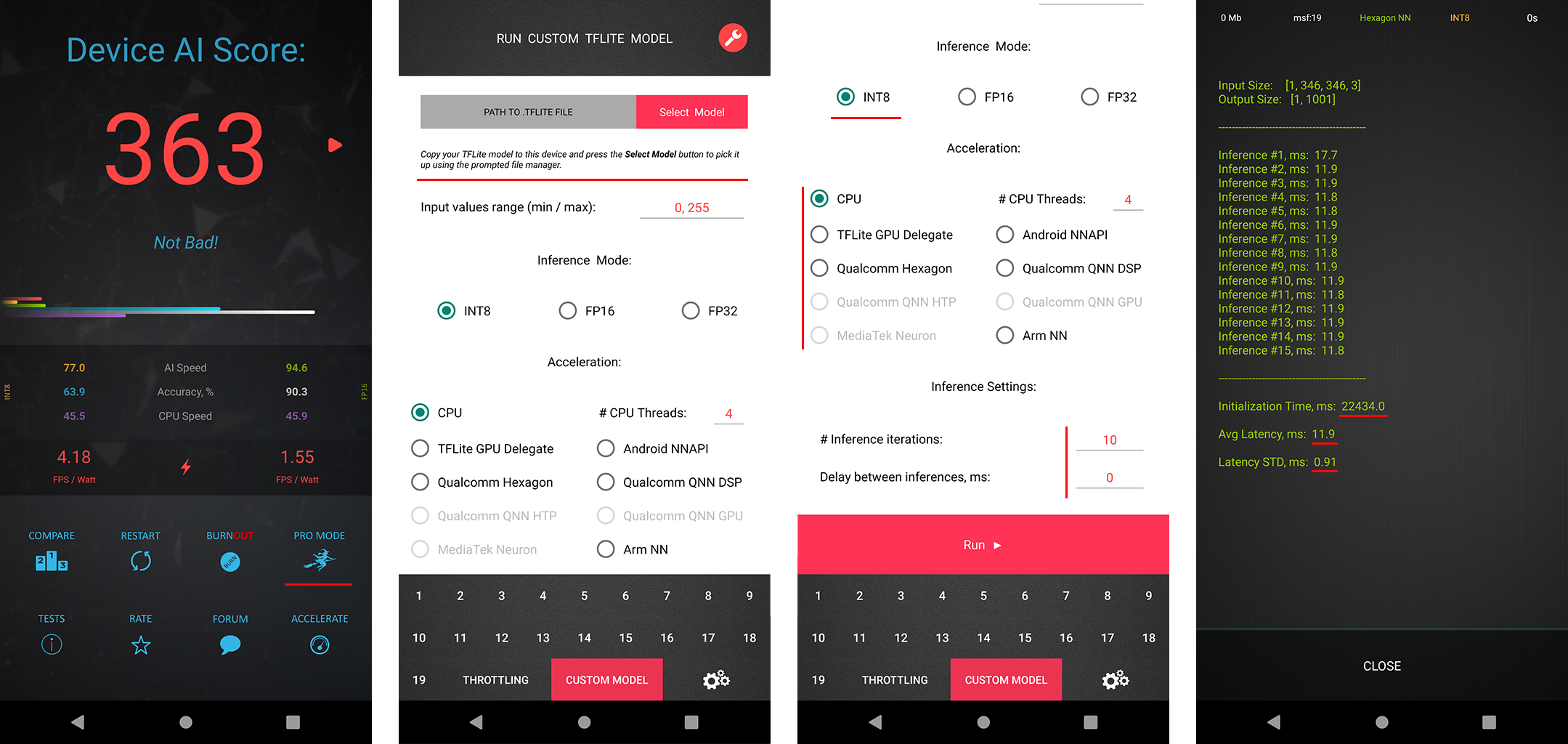}
}
\vspace{0.2cm}
\caption{Loading and running custom TensorFlow Lite models with AI Benchmark application. The currently supported acceleration options include Qualcomm QNN, Hexagon NN, MediaTek Neuron, TFLite GPU, Android NNAPI and ARM NN delegates as well as CPU inference (LiteRT / XNNPACK backends). The latest app version can be downloaded at \url{https://ai-benchmark.com/download}}
\label{fig:ai_benchmark_custom}
\end{figure*}

\section{Challenge}

To design an efficient and practical solution for mobile-related tasks, the following key components are essential:


\begin{enumerate}
\item A high-quality and large-scale dataset that can be used to train and evaluate the solution;
\item An efficient way to check the runtime and debug the model locally without any constraints;
\end{enumerate}

This challenge addresses the above issues. Real training data, tools, and runtime evaluation options provided to the challenge participants are described in the next sections.

\subsection{Dataset}

The participants were proposed to work with a real-world dataset consisting of noisy-clean images collected specifically for this challenge~\cite{ignatov2021fastDenoising}. For this, we used a recent Sony mobile camera sensor and captured photos in the burst mode: for each scene, 20 images were obtained and then averaged to get a clean photo. The illumination conditions varied from moderate to completely dark throughout the data collection process, the images were shot indoors and outdoors to get a variety of in-the-wild scenes with different noise patterns. The photos of more than 1000 different scenes were obtained and then checked manually to remove out-of-focus images and the ones where the built-in optical image stabilization caused misalignments. It should be additionally mentioned that we shot RAW photos using Android Camera2API~\cite{Camera2API2021} to avoid any effects from smartphones' integrated image denoising ISP modules, and the resulting images were then converted to RGB format using a stand-alone classical ISP system with disabled noise correction options. The resolution of the images was 4000$\times$3000 pixels, around 650 scenes were used for training the models, while the remaining photos were reserved for validation and testing. An example set of collected images is shown in Fig.~\ref{fig:example_photos}.

\subsection{Local Runtime Evaluation}

When developing AI solutions for mobile devices, it is vital to be able to test the designed models and debug all emerging issues locally on available devices. For this, the participants were provided with the \textit{AI Benchmark} application~\cite{ignatov2018ai,ignatov2019ai} that allows to load any custom TensorFlow Lite model and run it on any Android device with all supported acceleration options. This tool contains the latest versions of the \textit{Qualcomm QNN, MediaTek Neuron, Hexagon NN, TFLite GPU, Android NNAPI, Samsung ENN, Amlogic AML} and \textit{Arm NN} delegates, therefore supporting all current mobile platforms and providing the users with the ability to execute neural networks on smartphone NPUs, APUs, DSPs, GPUs and CPUs.

\smallskip

To load and run a custom TensorFlow Lite (LiteRT) model, one needs to follow the next steps:

\begin{enumerate}
\setlength\itemsep{0mm}
\item Download AI Benchmark from the official website\footnote{\url{https://ai-benchmark.com/download}} or from Google Play\footnote{\url{https://play.google.com/store/apps/details?id=org.benchmark.demo}} and run its standard tests.
\item After the end of the tests, enter the \textit{PRO Mode} and select the \textit{Custom Model} tab there.
\item Put the exported TFLite (LiteRT) model into the \textit{Download} folder of the device.
\item Select mode type \textit{(INT8, FP16, or FP32)}, the desired acceleration / inference options and run the model.
\end{enumerate}

\noindent These steps are also illustrated in Fig.~\ref{fig:ai_benchmark_custom}. The instructions on how to generate TFLite model from PyTorch, TensorFlow and Keras pre-trained models were provided in the MAI Workshop GitHub repository~\footnote{\url{https://github.com/aiff22/MAI-2025-Workshop}}.

\begin{table*}[t!]
\centering
\resizebox{\linewidth}{!}
{
\begin{tabular}{l|c|cc|cc|ccccc|c}
\hline
Team \, & \, Author \, & \, Framework \, & \, Model Size, \, & \, PSNR$\uparrow$ \, & \, SSIM$\uparrow$ \, & \multicolumn{2}{c}{\, Snapdragon 8 Elite Gen 5\,} & \multicolumn{2}{c}{\, Dimensity 9500\,} & Avg. GPU & \, Final Score \\
& & & \small{KB} & & & \, \small{GPU, ms $\downarrow$} \, & \, \small{NPU, ms $\downarrow$} \, &\, \small{GPU, ms $\downarrow$} \, & \, \small{NPU, ms $\downarrow$} \, &  \small{Runtime, ms $\downarrow$} & \\
\hline
\hline
TLG & perpetuity & PyTorch / Keras & 194 & 37.31 & 0.9129 & 28 & 46 & 50.7 & 18.7 & 38 & 101.1 \\
Fanis & nikolasent & PyTorch & 469 & 37.48 & 0.912 & 48.1 & 64.9 & 112 & \small{Failed} & 73 & 66.6 \\
TeamPAK & faraz2322 & PyTorch & 7885 & 37.58 & 0.9098 & 127 & 46.1 & 138 & 34 & 132 & 42.3 \\
Z6 & rlghksdbs & PyTorch & 97.3 & 36.96 & 0.9001 & 69.3 & 51.6 & 115 & 58.1 & 89 & 26.6 \\
IVCL & yoonsuk & PyTorch & 448 & 37.28 & 0.911 & 1961 & 552 & 720 & \small{Failed} & 1188 & 3.1 \\
wurbane & wurbane & PyTorch & 138948 & 35.22 & 0.8964 & 2292 & 633 & 2824 & 524 & 2544 & 0.1 \\
NormalVision & \_yang & PyTorch & 585 & 33.23 & 0.7767 & 945 & \small{Failed} & 1516 & \small{Failed} & 1197 & 0.01 \\
\hline
\end{tabular}
}
\vspace{2.6mm}
\caption{\small{Mobile AI 2026 Image Denoising challenge results and final rankings. During the runtime measurements, the models were performing denoising images of resolution 1920$\times$1088 pixels on the Snapdragon 8 Elite Gen 5 and Dimensity 9500 SoCs. Team \textit{TLG} is the challenge winner.}}
\label{tab:results}
\end{table*}

\subsection{Runtime Evaluation on the Target Platform}

In this challenge, we use two flagship mobile SoCs as our target evaluation platforms: \textit{Qualcomm Snapdragon 8 Elite Gen 5} and \textit{MediaTek Dimensity 9500} chipsets with the Adreno 840 and Arm Mali G1-Ultra MC12 GPUs, respectively, that can be found in numerous flagship smartphones. The models were parsed and accelerated using the TFLite GPU delegate~\cite{lee2019device} that demonstrated the best runtime results on both platforms. NPU runtimes provided for reference were obtained by running the models with the Qualcomm QNN and MediaTek Neuron TFLite delegates on the Snapdragon 8 Elite Gen 5 and the Dimensity 9500 SoCs. All final solutions were tested using the aforementioned AI Benchmark application.

\subsection{Challenge Phases}

The challenge consisted of the following phases:

\vspace{-0.8mm}
\begin{enumerate}
\item[I.] \textit{Development:} the participants get access to the data and AI Benchmark app, and are able to train the models and evaluate their runtime locally;
\item[II.] \textit{Validation:} the participants can upload their models to the remote server to check the fidelity scores on the validation dataset, and to compare their results on the validation leaderboard;
\item[III.] \textit{Testing:} the participants submit their final results, codes, TensorFlow Lite models, and factsheets.
\end{enumerate}
\vspace{-0.8mm}

\subsection{Scoring System}

All solutions were evaluated using the following metrics:

\begin{itemize}
\setlength\itemsep{-0.2mm}
\item Peak Signal-to-Noise Ratio (PSNR) measuring fidelity score,
\item Structural Similarity Index Measure (SSIM), a proxy for perceptual score,
\item The runtime on the target Snapdragon 8 Elite Gen 5 and Dimensity 9500 GPUs.
\end{itemize}

The score of each final submission was evaluated based on the next formula ($C$ is a constant normalization factor):

\smallskip
\begin{equation*}
\text{Final Score} \,=\, \frac{2^{2 \cdot \text{PSNR}}}{C \cdot \text{runtime}},
\end{equation*}
\smallskip

During the final challenge phase, the participants did not have access to the test dataset. Instead, they had to submit their final TensorFlow Lite models that were subsequently used by the challenge organizers to check both the runtime and the fidelity results of each submission under identical conditions. This approach solved all the issues related to model overfitting, reproducibility of the results, and consistency of the obtained runtime / accuracy values.

\section{Challenge Results}

In this challenge, 7 teams entered the final phase and submitted their results, TFLite models, codes, executables and factsheets. Table~\ref{tab:results} summarizes the final challenge results for both tracks and reports PSNR, SSIM and runtime numbers for the valid solutions on the final test dataset and on the target evaluation platform. The proposed methods are described in section~\ref{sec:solutions}, and the team members and affiliations are listed in Appendix~\ref{sec:apd:team}.

\subsection{Results and Discussion}

All submitted solutions demonstrated a very high efficiency: the majority of models are able to process Full HD (1920$\times$1088 px) image under 150 milliseconds on both target GPUs. This shows great progress in both model design as well as in mobile AI acceleration hardware that is currently able to provide decent speed-up for many common network architectures. Team \textit{TLG} is the challenge winner: solution proposed by this team was able to demonstrate over 35 and 19 FPS when running on the Adreno and Arm GPUs, respectively. It should be mentioned that this model is additionally compatible with NPUs from both Qualcomm and MediaTek, yielding an additional 2.5$\times$ runtime speed-up in the latter case. Similar numerical results were demonstrated by team \textit{Fanis}, which solution was slightly slower on both platforms. Team \textit{TeamPAK} achieved the highest fidelity score in this challenge. Despite showing worse latency on GPUs, this solution is compatible with both NPUs, where it can show a significant speed-up with a runtime of less than 50ms on Qualcomm and MediaTek NPUs. 

A notable progress that should be highlighted is the ease of model conversion to the target TFLite format. Table~\ref{tab:results} shows that the majority of solutions in this challenge were developed using the PyTorch ML framework. Due to the recently presented \textit{litert-torch} plugin~\footnote{\url{https://github.com/google-ai-edge/litert-torch}} developed by the LiteRT team, one can now directly convert the majority of PyTorch models to the TFLite format without the need for TensorFlow / Keras model re-implementation or use of ONNX as an intermediary format. This significantly simplifies the development process of mobile-friendly deep learning architectures for the majority of researchers using PyTorch as their main framework for AI model development and training.

\section{Challenge Methods}
\label{sec:solutions}

\noindent This section describes solutions submitted by all teams participating in the final stage of the MAI 2026 Image Denoising challenge.

\input{sec/01-tlg}

\input{sec/02-fanis}

\input{sec/03-pak}

\input{sec/04-z6}

\input{sec/05-ivcl}

\section{Additional Literature}

An overview of the past challenges on mobile-related tasks together with the proposed solutions can be found in the following papers:

\begin{itemize}
\item Image Super-Resolution:\, \cite{ignatov2018pirm,cai2019ntire,timofte2018ntire,lugmayr2020ntire}
\item Learned End-to-End ISP:\, \cite{ignatov2019aim,ignatov2020aim,wirzberger2022lan,ignatov2022microisp,ignatov2022pynetv2}
\item RGB Image Enhancement:\, \cite{ignatov2018pirm,ignatov2019ntire,ignatov2017dslr,ignatov2018wespe}
\item Bokeh Effect Rendering:\, \cite{ignatov2019aimBokeh,ignatov2020aimBokeh,ignatov2020rendering}
\item Image Denoising:\, \cite{abdelhamed2019ntire,abdelhamed2020ntire,flepp2024real}
\end{itemize}

\section*{Acknowledgements}

We thank AI Witchlabs, ETH Z\"urich, and the University of W\"urzburg (Computer Vision Lab), the organizers and sponsors of this Mobile AI 2026 challenge.

\appendix
\section{Teams and Affiliations}
\label{sec:apd:team}

\bigskip

\subsection*{Mobile AI 2026 Team}
\noindent\textit{\textbf{Title: }}\\ Mobile AI 2026 Image Denoising Challenge\\
\noindent\textit{\textbf{Members:}}\\ Andrey Ignatov$^{1,3}$ \textit{(andrey@vision.ee.ethz.ch)}, Georgy Perevozchikov$^2$ \textit{(georgii.perevozchikov@uni-wuerzburg. de)}, Radu Timofte$^{2,3}$ \textit{(radu.timofte@uni-wuerzburg.de)}\\
\noindent\textit{\textbf{Affiliations: }}\\
$^1$ Computer Vision Lab, ETH Z\"urich, Switzerland\\
$^2$ Computer Vision Lab, University of W\"urzburg, Germany\\
$^3$ AI Witchlabs, Switzerland\\

\subsection*{TLG}
\noindent\textit{\textbf{Members:}}\\ HaoDong Wu$^1$ \textit{(2991889017@qq.com)}, Desheng Cai$^1$\\
\noindent\textit{\textbf{Affiliations: }}\\
$^1$ Tianjin University of Technology, China\\

\subsection*{Fanis}
\noindent\textit{\textbf{Title:}}\\ FastSHADE: Fast Self-augmented Hierarchical Asymmetric Denoising for Efficient inference on mobile devices~\cite{falaleev2026fastshade}\\
\noindent\textit{\textbf{Members:}}\\ Nikolay Falaleev$^1$ \textit{(n.falaleev@fanis.ai)}\\
\noindent\textit{\textbf{Affiliations: }}\\
$^1$ Fanis Technologies Ltd., UK\\

\subsection*{TeamPAK}
\noindent\textit{\textbf{Title:}}\\ Real Image Denoising with Knowledge Distillation for High-Performance Mobile NPUs~\cite{kayani2026real}\\
\noindent\textit{\textbf{Members:}}\\ Faraz Kayani$^1$ \textit{(faraz.kayani2322@gmail.com)}, Sarmad Kayani$^1$, Asad Ahmed$^1$, Dmitry Ignatov$^1$\\
\noindent\textit{\textbf{Affiliations: }}\\
$^1$ Computer Vision Lab, CAIDAS \& IFI, University of Würzburg, Germany\\

\subsection*{Z6}
\noindent\textit{\textbf{Members:}}\\ Kihwan Yoon$^1$ \textit{(rlghksdbs@gmail.com)}, Ganzorig Gankhuyag$^2$\\
\noindent\textit{\textbf{Affiliations: }}\\
$^1$ BLUEDOT, Korea\\
$^2$ Korea Electronics Technology Institute (KETI), Korea\\

\subsection*{IVCL}
\noindent\textit{\textbf{Members:}}\\ Yoonsuk Kang \textit{(kangus214@sju.ac.kr)}, The Van Le \textit{(thele@sju.ac.kr)}, Jin Young Lee$^1$\textit{(jinyounglee@sju.ac.kr)}\\
\noindent\textit{\textbf{Affiliations: }}\\
$^1$ Intelligent Visual Computing Lab (IVCL), Sejong University, South Korea\\


\subsection*{Normal Vision}
\noindent\textit{\textbf{Members:}}\\ Sunzhe Yang$^1$ \textit{(52295901001@stu.ecnu.edu.cn)}, Qiuhai Yan$^1$, Kang Chen$^2$, Tingting Wang$^1$, Faming Fang$^1$\\
\noindent\textit{\textbf{Affiliations: }}\\
$^1$ East China Normal University, China\\
$^2$ Jiangxi Normal University, China\\

{\small
\bibliographystyle{ieeenat_fullname}
\bibliography{egbib}
}

\end{document}

%% file: sec/01-tlg.tex
\subsection{TLG}

\begin{figure}[h!]
\centering
\setlength{\tabcolsep}{1pt}
\resizebox{0.96\linewidth}{!}
{
\includegraphics[width=1.0\linewidth]{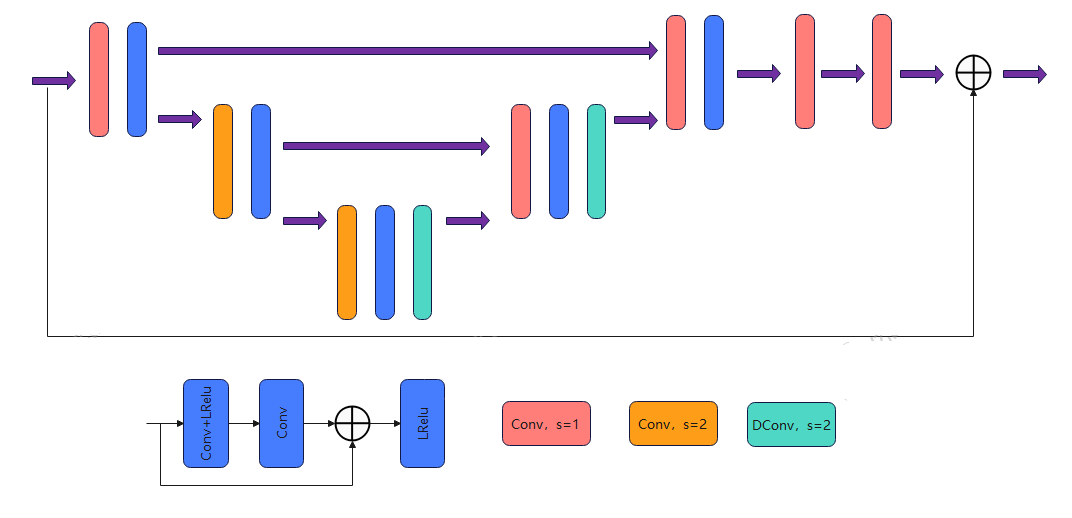}
}
\vspace{0.2cm}
\caption{DenoiseUnet architecture proposed by team TLG}
\label{fig:TLG}
\end{figure}

Team TLG proposed DenoiseUnet, a hierarchical encoder-decoder architecture illustrated in Fig.~\ref{fig:TLG}. The network employs a U-Net-like structure processing features across three spatial scales. The encoder path progressively reduces resolution using strided convolutions to capture multi-scale context. At each level, feature extraction is performed by residual blocks consisting of a convolution-LeakyReLU-convolution sequence, where the input is added back to the processed features before a final activation. The bottleneck at the lowest resolution level further refines the feature representation before the decoding stage. The decoder path restores the spatial dimensions using transposed convolutions for upsampling. To preserve high-frequency details and improve gradient flow, symmetric skip connections bridge corresponding encoder and decoder levels. Furthermore, the architecture implements global residual learning: a long skip connection propagates the original input directly to the end of the network, where it is summed with the output of the final decoder stage to produce the denoised image.

The model was trained for 5000 epochs using the $L_2$ loss function as the primary fidelity metric. The training process utilized the Adam optimizer with a batch size of 64 and input patches of size $256\times256$. The initial learning rate was set to $5\text{e-}4$ and was progressively reduced to $1\text{e-}8$ using a KneeLRScheduler.

%% file: sec/02-fanis.tex
\subsection{Fanis}

\begin{figure}[h!]
\centering
\setlength{\tabcolsep}{1pt}
\resizebox{0.96\linewidth}{!}
{
\includegraphics[width=1.0\linewidth]{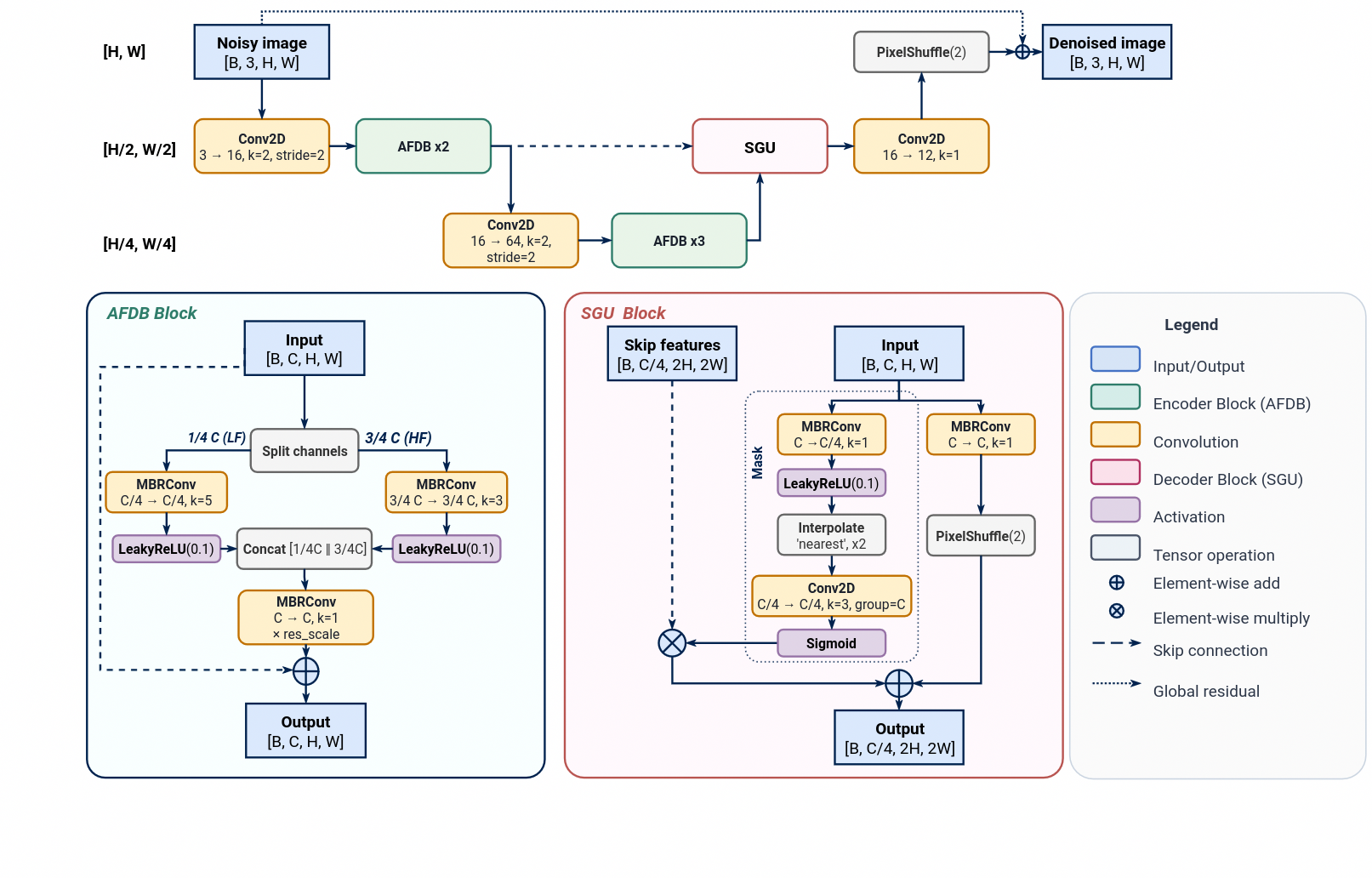}
}
\vspace{0.2cm}
\caption{FastSHADE~\cite{falaleev2026fastshade} architecture proposed by team Fanis}
\label{fig:Fanis}
\end{figure}

Team Fanis developed FastSHADE~\cite{falaleev2026fastshade}, a U-Net-style architecture illustrated in Fig.~\ref{fig:Fanis}. The network computes representations at two spatial scales, utilizing learnable $2\times2$ convolutions with a stride of 2 for spatial downsampling. The core feature extraction is driven by Asymmetric Frequency Denoising Blocks (AFDB), which split incoming feature maps into two parallel channel paths. A low-frequency path processes 25\% of the channels using a large-kernel $5\times5$ convolution to extract broad structural context, while a high-frequency path processes the remaining 75\% using a $3\times3$ convolution for rapid noise suppression. These branches are subsequently concatenated, fused via a $1\times1$ convolution, and merged with a residual connection. Feature decoding is performed using a Spatially Gated Upsampler (SGU) and PixelShuffle operations. The SGU dynamically modulates high-resolution skip connections with a generated spatial mask, allowing essential structural details to propagate efficiently. To ensure real-time mobile performance, all blocks are constructed using Multi-Branch Re-parameterized Convolutions (MBRConv)~\cite{yan2025mobileie}, which operate as complex multi-branch modules during training but are collapsed into efficient single dense convolutions for inference.

The model was trained using the provided Challenge data augmented with the SIDD-Medium~\cite{abdelhamed2018high} and MIDD~\cite{flepp2024real} datasets. Training followed a two-stage methodology. The first stage trained the model from scratch for 80 epochs using Charbonnier loss, the AdamW optimizer, a batch size of 32, and $256\times256$ input patches. A cosine annealing scheduler decayed the learning rate from $1\text{e-}3$ to $1\text{e-}6$. The second stage fine-tuned the model for 20 epochs using a PSNR-based loss function, larger $512\times512$ patches, a batch size of 20, and an initial learning rate of $1\text{e-}4$. Data augmentation included geometric transformations alongside a stochastic noise self-augmentation strategy, which synthesized new inputs by blending ground-truth noise with network-predicted noise from other batch samples. Training stability was maintained using Exponential Moving Average (EMA) with a 0.999 decay rate. Finally, a Straight-Through Estimator (STE) was employed during training to apply gradient-preserving rounding, ensuring robust directly-quantized integer predictions at the output stage.

%% file: sec/03-pak.tex
\subsection{TeamPAK}

\begin{figure}[h!]
\centering
\setlength{\tabcolsep}{1pt}
\resizebox{0.96\linewidth}{!}
{
\includegraphics[width=1.0\linewidth]{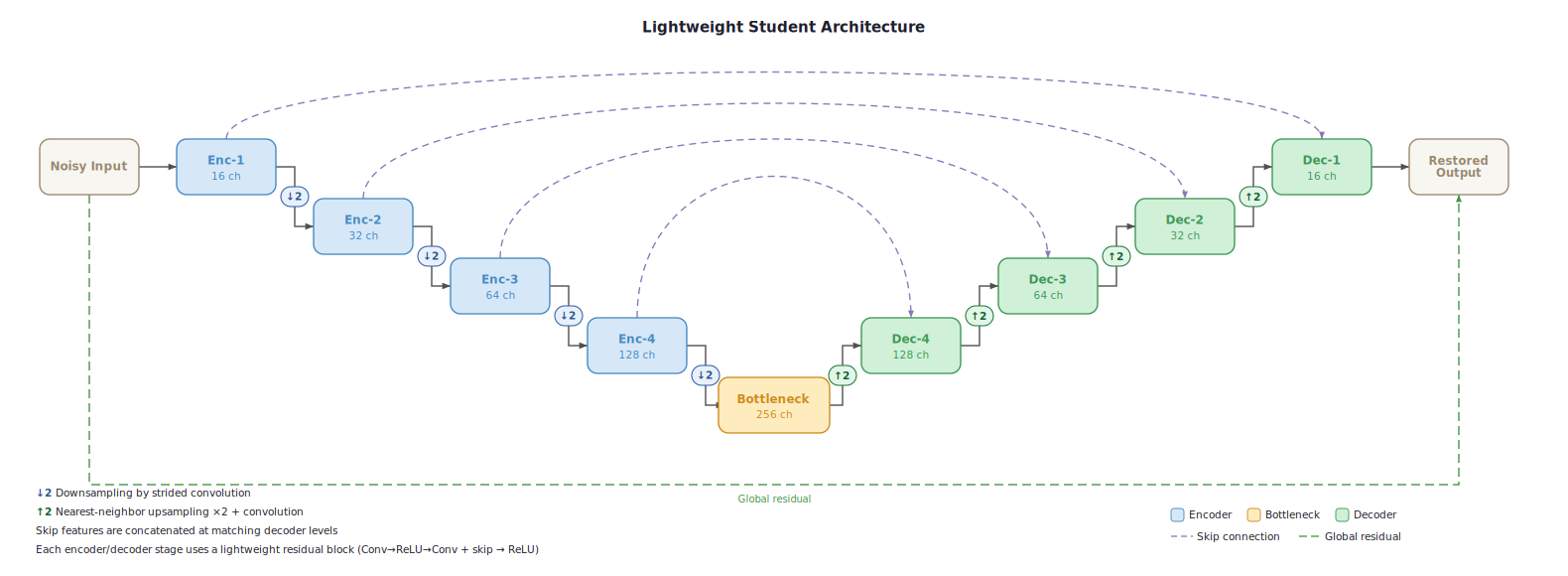}
}
\vspace{0.2cm}
\caption{Lightweight student architecture proposed by team TeamPAK}
\label{fig:TeamPAK}
\end{figure}

Team TeamPAK proposed a lightweight U-Net variant architecture, illustrated in Fig.~\ref{fig:TeamPAK}, optimized for mobile GPU delegates~\cite{kayani2026real}. The network employs a base filter count of 16, which scales up to 256 at the bottleneck. The core feature extraction units are Lite Denoising Blocks. Each block utilizes a sequence of a $3\times3$ convolution, a ReLU activation, and another $3\times3$ convolution, featuring an internal channel bottleneck that halves and then restores the number of channels ($f \rightarrow f/2 \rightarrow f$). An identity skip connection bypasses these convolutions and is added to the output before a final ReLU activation. In the decoder path, the model uses nearest-neighbour upsampling with a scale factor of 2 instead of transposed convolutions to maintain a low parameter footprint. Features from the encoder are concatenated via skip connections at corresponding decoder levels. Furthermore, a global identity residual connection links the input directly to the network's output. ReLU activations are utilized exclusively throughout the network to ensure broad hardware compatibility. To facilitate reproducibility, LiteDenoiseNet is publicly released as part of the LEMUR NN Dataset~\cite{ABrain.NN-Dataset, ABrain.LEMUR2}.

The training process used a synchronized Knowledge Distillation workflow~\cite{hinton2015distilling}, with a high-capacity teacher model guiding the lightweight student. The student model was initialized with a warm start from a prior best-performing checkpoint. Training was conducted for 200 epochs using the Adam optimizer with an initial learning rate of $1\text{e-}4$ and a weight decay of $1\text{e-}4$. A cosine annealing schedule smoothly decayed the learning rate to $1\text{e-}5$ by the end of training. The batch size was varied across runs, utilizing sizes of 8, 16, and 24. To ensure convolution kernels possessed full spatial context, input patches were initially sampled at a size of $258\times258$ and subsequently center-cropped to $256\times256$. A heavily weighted distillation loss function was employed to force the student to prioritize the teacher's clean reconstructions over potentially noisy ground truth labels. The total loss combined Mean Squared Error (MSE) against the teacher's output (weighted at 900), MSE against the ground truth (weighted at 100), and $L_1$ loss against the ground truth (weighted at 50). Additionally, gradient clipping with a maximum norm of 0.1 was applied at each step to maintain stability under the high distillation weight.

%% file: sec/04-z6.tex
\subsection{Z6}

\begin{figure}[h!]
\centering
\setlength{\tabcolsep}{1pt}
\resizebox{0.96\linewidth}{!}
{
\includegraphics[width=1.0\linewidth]{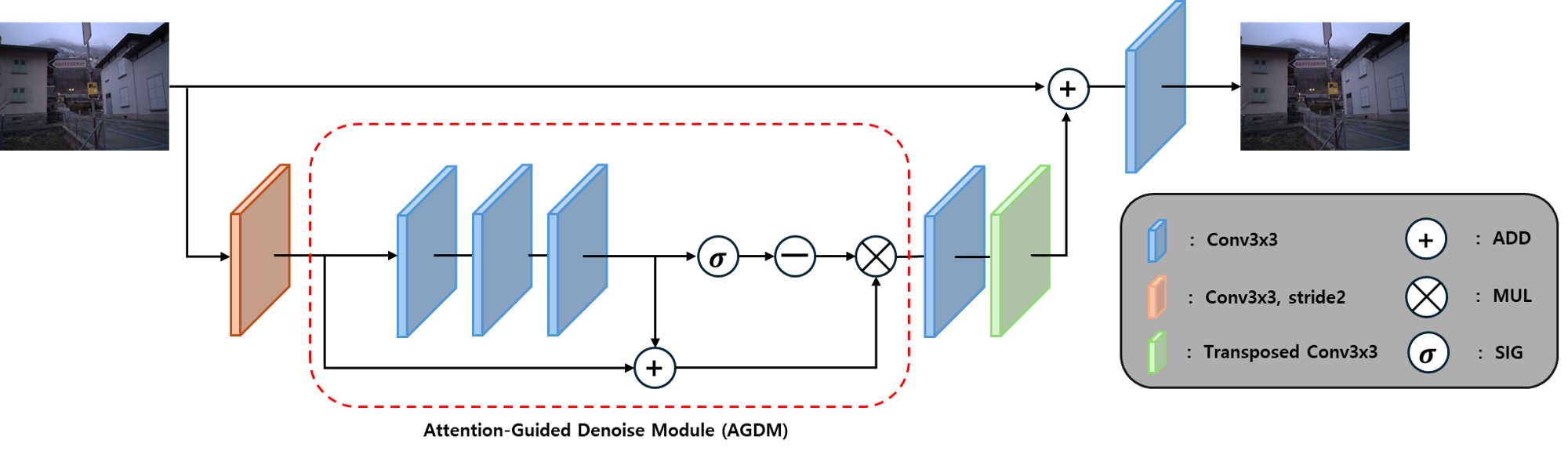}
}
\vspace{0.2cm}
\caption{AGDNet architecture proposed by team Z6}
\label{fig:Z6}
\end{figure}

Team Z6 proposed the Attention-Guided Deblur Network (AGDNet), illustrated in Fig.~\ref{fig:Z6}; based on GFAB~\cite{ren2025tenth}. To optimize both speed and accuracy, the network first applies a $3\times3$ convolution with a stride of 2 for spatial downsampling. The core feature extraction is driven by the Attention-Guided Denoise Module (AGDM). This module combines two Reparameterized Blocks~\cite{yoon2024casr, gankhuyag2023lightweight} with a parameter-free attention mechanism, where the original activation is replaced with ReLU. During training, overparameterized blocks are selectively utilized only where input and output shapes match; these are subsequently fused into single $3\times3$ convolutional layers to accelerate inference. The network maintains a fixed width of 24 feature channels throughout its hidden layers. After the AGDM processing, a transposed $3\times3$ convolution with a stride of 2 is employed for spatial upsampling. Finally, a global skip connection bypasses the main network, adding the original input image directly to the upsampled features to generate the final denoised output.

The model was trained from scratch exclusively on the MAI2021 denoising dataset. Training spanned 200 epochs using the $L_1$ loss function. The optimization process utilized the Adam optimizer with a mini-batch size of 8 and randomly cropped input patches of size $384\times384$. The initial learning rate was set to $5\text{e-}3$ and was controlled by a cosine warm-up scheduler with a 0.1\% warm-up ratio.

%% file: sec/05-ivcl.tex
\subsection{IVCL}

\begin{figure}[h!]
\centering
\setlength{\tabcolsep}{1pt}
\resizebox{0.96\linewidth}{!}
{
\includegraphics[width=1.0\linewidth]{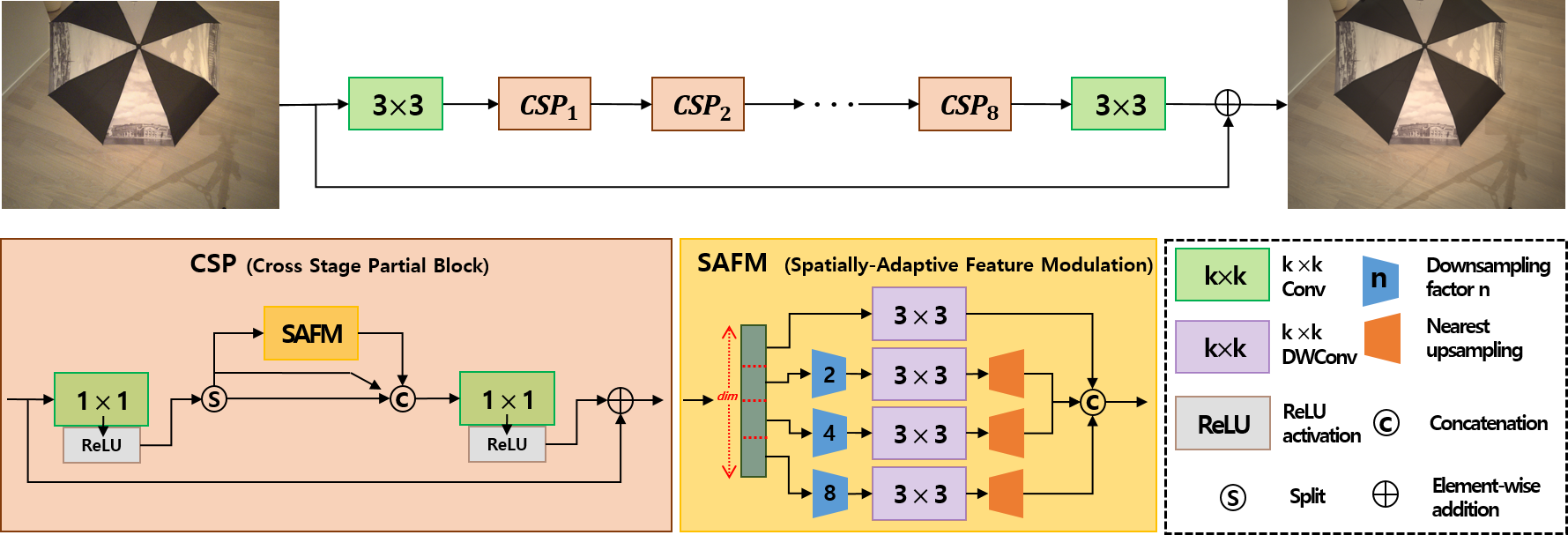}
}
\vspace{0.2cm}
\caption{CSPID architecture proposed by team IVCL}
\label{fig:IVCL}
\end{figure}

Team IVCL proposed the Cross-Stage Partial ImageDenoising (CSPID) network, illustrated in Fig.~\ref{fig:IVCL}, based on CSPSR ~\cite{telili2025360}. The architecture employs a global residual learning strategy where the original input image is added directly to the final predicted residual. The network pipeline begins with a $3\times3$ convolution for shallow feature extraction. The core deep feature extraction is performed by a sequence of 8 Cross-Stage Partial (CSP) blocks~\cite{wang2020cspnet}, maintaining 64 intermediate channels throughout. To reduce computational complexity without sacrificing performance, the team integrated Spatially-Adaptive Feature Modulation (SAFM)~\cite{sun2023spatially} into the CSP structure. Within each CSP block, incoming features are split: one path undergoes processing via $1\times1$ convolutions and the SAFM module, while the other acts as a bypass. The paths are then concatenated, compressed via a $1\times1$ convolution, and added back to the block's input via a local residual connection. The internal SAFM module captures multi-scale context by routing features through parallel branches with downsampling factors of 2, 4, and 8, utilizing $3\times3$ depthwise and standard convolutions before applying nearest-neighbor upsampling. Finally, a $3\times3$ convolution acts as the reconstruction module to produce the final denoised features.

The model was trained on the Mobile AI 2021 denoising dataset. To prepare the data, 30 overlapping $256\times256$ patches were extracted from each high-resolution training image. During the training phase, the model ingested patches of size $128\times128$, obtained through random cropping, and augmented with random flipping and rotation. The training process ran for 1,000,000 iterations using a batch size of 32. The network was optimized using the Charbonnier loss function with a learning rate set to $1\text{e-}4$.